\documentclass{article}

\usepackage[nonatbib]{nips_2017}

\usepackage[utf8]{inputenc} 
\usepackage[T1]{fontenc}    
\usepackage{hyperref}       
\usepackage{url}            
\usepackage{booktabs}       
\usepackage{amsfonts}       
\usepackage{nicefrac}       
\usepackage{microtype}      
\usepackage{graphicx}
\usepackage{amsmath,amssymb} 
\usepackage{color}
\usepackage{hyphenat}
\usepackage{subfigure}
\long\def\invis#1{}
\usepackage{multirow}
\usepackage{tipa}
\usepackage{caption}
\title{An Ensemble of Deep Convolutional Neural Networks for Alzheimer’s Disease Detection and Classification}

\author{
  Jyoti Islam \\
  Department of Computer Science\\
  Georgia State University\\
  \texttt{jislam2@student.gsu.edu} \\
   \And
Yanqing Zhang \\
Department of Computer Science \\
Georgia State University \\
   \texttt{yzhang@gsu.edu} \\
}

\begin{document}

\maketitle

\begin{abstract}
Alzheimer's Disease destroys brain cells causing people to lose their memory, mental functions and ability to continue daily activities. It is a severe neurological brain disorder which is not curable, but earlier detection of Alzheimer's Disease can help for proper treatment and to prevent brain tissue damage. Detection and classification of Alzheimer's Disease (AD) is challenging because sometimes the signs that distinguish Alzheimer's Disease MRI data can be found in normal healthy brain MRI data of older people. Moreover, there are relatively small amount of dataset available to train the automated Alzheimer's Disease detection and classification model. In this paper, we present a novel  Alzheimer's Disease detection and classification model using brain MRI data analysis. We develop an ensemble of deep convolutional neural networks and demonstrate superior performance on the Open Access Series of Imaging Studies (OASIS) dataset.

\end{abstract}
\section{Introduction}

\begin{figure}[h]
\begin{center}
\leavevmode
\begin{tabular}{cccc}
\subfigure[]{\includegraphics[width=.20\linewidth, height=0.75in]{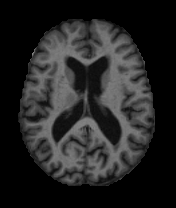}\label{fig:resp1}}&
\subfigure[]{\includegraphics[width=.20\linewidth,height=0.75in]{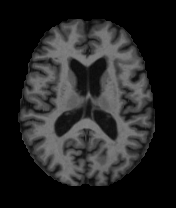}\label{fig:resp2}}
\subfigure[]{\includegraphics[width=.20\linewidth,height=0.75in]{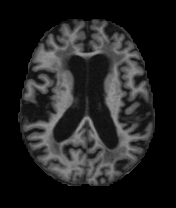}\label{fig:resp3}}&
\subfigure[]{\includegraphics[width=.20\linewidth,height=0.75in]{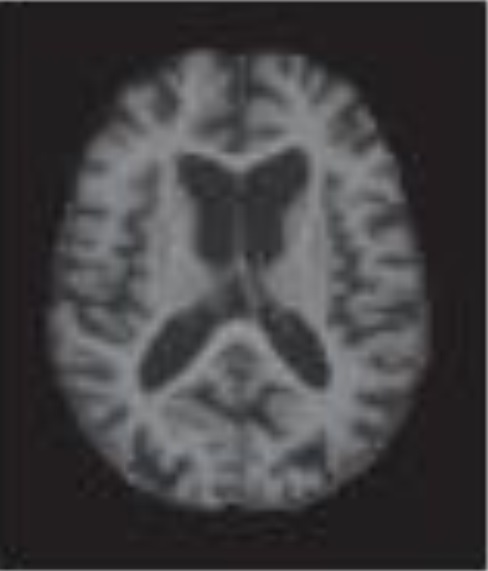}\label{fig:resp4}}
\end{tabular}
\end{center}
\caption{Example of different brain MRI images presenting different AD stage. \subref{fig:resp1} Nondemented; \subref{fig:resp2} very mild dementia ;
\subref{fig:resp3} mild dementia;
 \subref{fig:resp4}  moderate dementia.}
\label{fig_resp}
\end{figure}

There are three major stages in Alzheimer's Disease - very mild, mild and moderate.  Detection of Alzheimer's Disease (AD) is still not accurate until a patient reaches moderate AD. But early detection and classification of AD are critical for proper treatment and preventing brain tissue damage. Alzheimer's disease has a certain progressive pattern of brain tissue damage. It shrinks the hippocampus and cerebral cortex of the brain and enlarges the ventricles \cite{sarraf2016deepad}. Hippocampus is the responsible part of the brain for episodic and spatial memory. It also works as a relay structure between our body and brain. The reduction in hippocampus causes cell loss and damage specifically to synapses and neuron ends. So neurons cannot communicate anymore via synapses. As a result, brain regions related to remembering (short term memory), thinking, planning, and judgment are affected \cite{sarraf2016deepad}. The degenerated brain cells have low intensity in MRI images \cite{warsi2012fractal}. Figure~\ref{fig_resp} shows some brain MRI images presenting different AD stage.

Machine learning studies using neuroimaging data for developing diagnostic tools helped a lot for automated brain MRI segmentation and classification. Most of them use handcrafted feature generation and extraction from the MRI data. After that, the features are fed into machine learning models such as Support Vector Machine, Logistic regression model, etc. These multi-step architectures are complex and highly dependent on human experts. Besides, the size of datasets for neuroimaging studies is small. While image classification datasets used for object detection and classification has millions of images (for example, ImageNet database \cite{ILSVRC15}), neuroimaging datasets typically have less than 1000 images. But to develop robust neural networks we need a lot of images. Because of the scarcity of large image database, it is important to develop models that can learn useful features from the small dataset. For our research work, we developed an ensemble of deep convolutional neural networks for Alzheimer's Disease detection and classification using a small dataset (OASIS database \cite{marcus2007open}).

\begin{figure}[h]
\centering
\includegraphics[width=4.00in, height=1.65in]{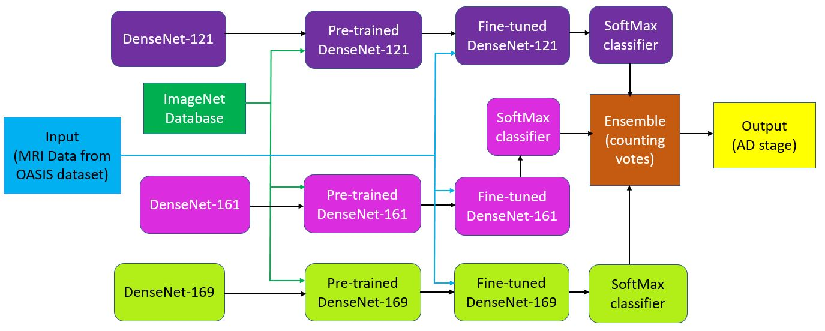}
\caption{Block diagram of proposed Alzheimer's disease detection and classification framework. }
\label{fig_model}
\end{figure}
\section{Related Work}

Some remarkable research work have been done for automated Alzheimer's Disease detection and classification. But majority of them have used ADNI dataset \cite{jack2008alzheimer}. For example, Aversen et al.\cite{arvesen2015automatic} used dimensional reduction and variations methods, Brosch et al.\cite{brosch2013manifold} developed a deep belief network model, Gupta et al.\cite{gupta2013natural} developed a sparse autoencoder model and Payan et al. \cite{payan2015predicting} combined sparse autoencoders and 3D CNN model for AD detection and classification using ADNI dataset \cite{jack2008alzheimer}.
Amulya et al. \cite{er2017classification} used Gray-Level Co-occurrence Matrix (GLCM) method  for Alzheimer's Disease classification using OASIS database \cite{marcus2007open}. In our previous work \cite{Islam2017}, we developed a very deep neural network inspired by inception-v4 model \cite{Szegedy} for Alzheimer's Disease detection. We will use OASIS dataset to evaluate our proposed model to check whether the proposed model performs well for small medical image dataset or not.

\section{Proposed Network Architecture}
In this section, the proposed Alzheimer's disease detection and classification framework would be presented. The proposed model is shown in Figure~\ref{fig_model}. Our model is an ensemble of three DenseNet \cite{huang2017densely} styled model - DenseNet-121, DenseNet-161, and DenseNet-169. Here the numbers 121, 161 and 169 denote the depth of the models. For example, 121 comes from the computation: 5+(6+12+24+16)*2=121, where 5 is (conv,pooling)+3 transition layers+ classification layer. We multiply with 2 because each dense block has 2 layers (1x1 conv and 3x3 conv). In a DenseNet \cite{huang2017densely} styled model, each layer has direct connections to all subsequent layers. So, each layer receives the feature-maps from all preceding layers. The feature-maps work as global state of the network, where each layer can add its own feature-map. The global state can be accessed from any part of the network and how much each layer can contribute to is decided by the growth rate of the network \cite{huang2017densely}.

For each MRI data, we have created patches from three physical planes of imaging: Axial or horizontal plane, Coronal or frontal plane, and Sagittal or median plane. These patches are feed to the proposed network as input. We applied transfer learning \cite{Islam2016} and the DenseNet-121, DenseNet-161, and DenseNet-169 models have been pre-trained with ImageNet dataset \cite{ILSVRC15}. Each of the softmax layer has four different output classes: nondemented, very mild, mild and moderate AD. The individual models take an MRI image as input and generates its learned representation. Based on this feature representation, the input MRI image is classified to any of the four output classes. To measure the loss of each of these models, we have used cross  entropy. The Softmax layer takes the learned representation, f\textsubscript{i} and interprets it to the output class. A probability score, p\textsubscript{i} is also assigned for the output class. If we define the number of Alzheimer's disease stages as m, then we get
\[ p_{i} = \frac{exp(f{_i})}{\sum_{i}{} exp(f{_i})}, i=1,...,m \] and \[ L = -\sum_{i}{} t_i log (p_i), \]
where L is the loss of cross entropy of the network. Back propagation is used to calculate the gradients of the network. If the ground truth of an MRI image is denoted as t\textsubscript{i}, then,
\[ \frac{ \partial L}{ \partial f_i} = p_i - t_i \]

The output classification label of the three individual model are ensembled together using voting technique. Each classifier ``votes'' for a particular class, and the class with the majority votes would be assigned as the label for the input MRI data. Since the dataset is small, 5-fold cross validation is performed on the dataset. For each fold, We have used 70\% as training data, 10\% as validation data and 20\% as test data. The individual models are optimized with the Stochastic Gradient Descent (SGD) algorithm and early-stopping is used for regularization.

\section{Experiments}

\subsection{Dataset}
\begin{figure}[h]
\centering
\includegraphics[width=3.25in, height=0.75in]{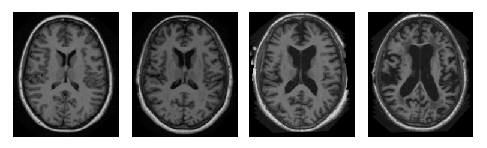}
\caption{Sample images from OASIS dataset \cite{marcus2007open}. }
\label{fig_sampleimages}
\end{figure}

OASIS dataset \cite{marcus2007open} have 416 subjects aged 18 to 96, and for each of them, 3 or 4 T1-weighted MRI scans are available. 100 of the patients having age over 60 are included in the dataset with very mild to moderate AD. Figure~\ref{fig_sampleimages} shows some sample brain MRI images from OASIS dataset. Since the dataset is small, 5-fold cross validation is performed on the dataset. 
\begin{table}[h]
\centering
\caption{Performance of the DenseNet-121 model on the OASIS dataset \cite{marcus2007open}}
\label{table_one}
\begin{tabular}{|l|c|c|c|c|}
\hline
Class & precision & recall & f1--score & support \\ \hline
non-demented  &  0.99     &   0.99  &   0.99  &   73  \\ \hline
very mild  &  0.75     &   0.50  &   0.60  &   6  \\ \hline
mild  &  0.62    &   0.71  &   0.67  &   7  \\ \hline
moderate  &  0.33     &   0.50  &   0.40  &   2  \\ \hline
avg/total  &  0.93     &   0.92  &   0.92  &   88  \\ \hline 
\end{tabular}
\end{table}

\begin{table}[h]
\centering
\caption{Performance of the DenseNet-161 model on the OASIS dataset \cite{marcus2007open}}
\label{table_two}
\begin{tabular}{|l|c|c|c|c|}
\hline
Class & precision & recall & f1--score & support \\ \hline
non-demented  &  0.88     &   0.95  &   0.91  &   73  \\ \hline
very mild  &  0.00     &   0.00  &   0.00  &   6  \\ \hline
mild  &  0.25     &   0.29 &   0.27  &   7  \\ \hline
moderate  &  0.00     &   0.00  &   0.00  &   2  \\ \hline
avg/total  &  0.75     &   0.81  &   0.78  &   88  \\ \hline 
\end{tabular}
\end{table}

\subsection{Results}

\begin{table}[h]
\centering
\caption{Performance of the DenseNet-169 model on the OASIS dataset \cite{marcus2007open}}
\label{table_three}
\begin{tabular}{|l|c|c|c|c|}
\hline
Class  & precision & recall & f1--score & support \\ \hline
non-demented  &  0.99     &   0.96  &   0.97  &   73  \\ \hline
very mild  &  0.50     &   0.33  &   0.40  &   6  \\ \hline
mild &  0.45     &   0.71  &   0.56  &   7  \\ \hline
moderate  &  0.50     &   0.50  &   0.50  &   2  \\ \hline
avg/total  &  0.90     &   0.89  &   0.89  &   88  \\ \hline 
\end{tabular}
\end{table}

\begin{table}[h]
\centering
\caption{Performance of the proposed ensembled model on the OASIS dataset \cite{marcus2007open}}
\label{table_proposed}
\begin{tabular}{|l|c|c|c|c|}
\hline
Class  & precision & recall & f1--score & support \\ \hline
non-demented  &  0.97     &   1.00  &   0.99  &   73  \\ \hline
very mild  &  1.00     &   0.33  &   0.50  &   6  \\ \hline
mild  &  0.67     &   0.86  &   0.75  &   7  \\ \hline
moderate  &  0.50     &   0.50  &   0.50  &   2  \\ \hline
avg/total  &  0.94     &   0.93  &   0.92  &   88  \\ \hline 
\end{tabular}
\end{table}

Table~\ref{table_one}, Table~\ref{table_two}, and Table~\ref{table_three} shows the individual performance of DenseNet-121, DenseNet-161, and DenseNet-169 model respectively. Table~\ref{table_proposed} shows the performance of our proposed ensembled model on the OASIS dataset \cite{marcus2007open} which is better than each of the individual DenseNet-121, DenseNet-161, and DenseNet-169 model. The accuracy of the proposed model is 93.18\% with 93\% precision, 92\% recall and 92\% f1--score. The performance comparison of classification results of the proposed ensembled model and the previous state--of--the--art is presented in Table~\ref{table_compare}. As we can see, proposed ensembled model achieves encouraging performance and outperforms the previous state-of the art. The reason behind the success of the proposed model with such small dataset is due to using the class weights in the training process.

\section{Conclusion}
We have provided an one step analysis for the 3D brain MRI data for Alzheimer's Disease detection and classification. The distinction between very mild and mild class would help to identify the current stage for early stage Alzheimer's Disease patient. Our system could be used for clinical decision making process to detect and classify different stages of Alzheimer's Disease. Though the proposed model has only been tested on Alzheimer's Disease data, we believe it can be used for other classification problems in the medical domain. Moreover, the proposed ensembled approach can be used for applying CNN into other domains with limited dataset.

\begin{table}[h]
\centering
\caption{Comparison of classification results on the OASIS dataset \cite{marcus2007open}}
\label{table_compare}
\begin{tabular}{|l|c|c|c|c|}
\hline
Methods & accuracy & precision & recall & f1--score \\ \hline
GLCM\cite{er2017classification}   &  75.71\%     &   63.84\%  & 100\% & 77.92\%\\ \hline
Proposed ensembled model  & 93.18\%      & 94\%   & 93\% & 92\% \\ \hline
\end{tabular}
\end{table}

\clearpage


\end{document}